\pdfoutput=1
\documentclass[11pt]{article}

\usepackage[]{acl}

\usepackage{times}
\usepackage{latexsym}
\usepackage{booktabs}
\usepackage{bm}
\usepackage{pifont}
\usepackage{multirow}
\usepackage{graphicx}
\usepackage{amsmath}
\usepackage{amssymb}
\usepackage{comment}
\usepackage{mathtools}
\usepackage{enumitem}
\usepackage{xspace}
\usepackage{tikz}
\usepackage{color, colortbl}

\setitemize{noitemsep,topsep=0pt,parsep=0pt,partopsep=0pt}
\usepackage[T1]{fontenc}
\usepackage[utf8]{inputenc}

\usepackage{microtype}

\usepackage{titlesec}
\titlespacing*{\section}{0pt}{1.4ex plus 0.6ex minus 0.3ex}{0.8ex plus 0.5ex minus 0.2ex}
\titlespacing*{\subsection}{0pt}{0.8ex plus 0.5ex minus 0.2ex}{0.5ex plus 0.5ex minus 0.2ex}

\makeatletter
\renewcommand{\paragraph}{%
  \@startsection{paragraph}{4}%
  {\z@}{.5ex \@plus .5ex \@minus .2ex}{-1em}%
  {\normalfont\normalsize\bfseries}%
}
\makeatother

\newcommand{\cmark}{\text{\ding{51}}}
\newcommand{\xmark}{\text{\ding{55}}}
\def\tinycol{\hskip 5pt}
\def\thincol{\hskip 2pt}

\definecolor{myorange}{RGB}{237, 125, 49}
\definecolor{myblue}{RGB}{68, 114, 196}
\newcommand{\correct}[1]{\textcolor{myblue}{#1}}
\newcommand{\wrong}[1]{\textcolor{myorange}{#1}}

\newcommand{\compacteq}{\setlength{\belowdisplayskip}{5pt}\setlength{\belowdisplayshortskip}{3pt}\setlength{\abovedisplayskip}{5pt}\setlength{\abovedisplayshortskip}{3pt}}

\newcommand{\tworow}[2]{\begin{matrix*}[l]
#1 \\
#2
\end{matrix*}}

\newcommand{\sparql}{\scalebox{0.6}{$\square$}\xspace}
\newcommand{\nl}{\tikz[baseline=-0.5ex]\draw[black,fill=black,radius=1.5pt] (0,0) circle;\xspace}
\newcommand{\concat}{\tikz[baseline=-0.5ex]\draw[black,fill=white,radius=1.5pt] (0,0) circle;\xspace}
\newcommand{\hlcell}{\cellcolor{gray!20}}

\newcommand\hfilll{\hspace{0pt plus 1filll}}
\newcommand{\toright}[1]{{\hfilll #1}}

\newcommand{\veryshortarrow}[1][4pt]{\mathrel{%
   \hbox{\rule[\dimexpr\fontdimen22\textfont2-.2pt\relax]{#1}{.4pt}}%
   \mkern-4mu\hbox{\usefont{U}{lasy}{m}{n}\symbol{41}}}}

\def\smallcol{\hskip 6pt}
\def\tinycol{\hskip 2pt}

\def\openoracle{oracle-book\xspace}

\title{Understanding and Improving Zero-shot Multi-hop Reasoning \\in Generative Question Answering}

\author{
Zhengbao Jiang$^\dag$,\quad Jun Araki$^\ddag$,\quad Haibo Ding$^\ddag$\thanks{~~Haibo Ding is now at Amazon.},\quad Graham Neubig$^\dag$ \\
  $^\dag$Languages Technologies Institute, Carnegie Mellon University \\
  $^\ddag$Bosch Research \\
\texttt{\{zhengbaj,gneubig\}@cs.cmu.edu} \\
\texttt{\{jun.araki,haibo.ding\}@us.bosch.com}
}

\begin{document}
\maketitle
\begin{abstract}
Generative question answering (QA) models generate answers to questions either solely based on the parameters of the model (the \emph{closed-book} setting) or additionally retrieving relevant evidence (the \emph{open-book} setting).
Generative QA models can answer some relatively complex questions, but the mechanism through which they do so is still poorly understood.
We perform several studies aimed at better understanding the multi-hop reasoning capabilities of generative QA models.
First, we decompose multi-hop questions into multiple corresponding single-hop questions, and find marked inconsistency in QA models' answers on these pairs of ostensibly identical question chains.
Second, we find that models lack zero-shot multi-hop reasoning ability: when trained only on single-hop questions, models generalize poorly to multi-hop questions.
Finally, we demonstrate that it is possible to improve models' zero-shot multi-hop reasoning capacity through two methods that approximate real multi-hop natural language (NL) questions by training on either concatenation of single-hop questions or logical forms (SPARQL).
In sum, these results demonstrate that multi-hop reasoning does not emerge naturally in generative QA models, but can be encouraged by advances in training or modeling techniques.%
\footnote{Code is available at \url{https://github.com/jzbjyb/multihop}.}
\end{abstract}

\section{Introduction}

\begin{figure}[tb]
\includegraphics[width=0.9\columnwidth, clip, keepaspectratio]{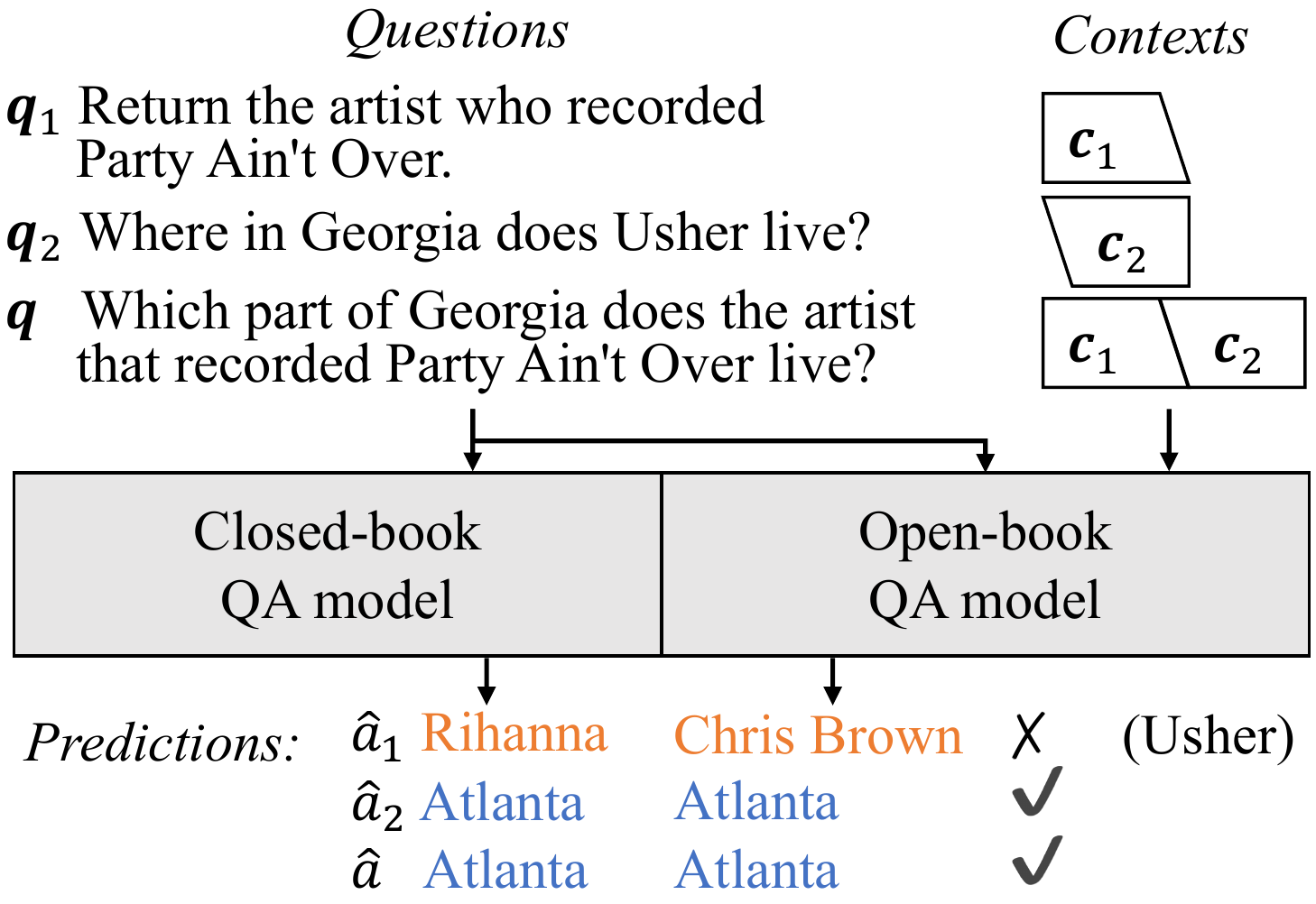}
\centering
\caption{Probing generative closed- and open-book QA models with both multi-hop ($\bm{q}$) and their component single-hop questions ($\bm{q}_1, \bm{q}_2$).}
\label{fig:thumbnail}
\end{figure}

Empowered by large-scale pre-trained language models (LMs) \cite{devlin-2018-bert,liu-2019-roberta,lewis-2020-bart,raffel-2020-t5}, recent years have seen much progress on \emph{generative question answering} (QA), where LMs generate answers given questions in an end-to-end fashion.
While most works only demonstrate the performance of such generative QA models on simple questions \cite{joshi-2017-triviaqa,kwiatkowski-2019-nq}, there has been some indication that these models can also answer complex questions that theoretically require multi-hop reasoning \cite{xiong-2020-mdr}, sometimes to an impressive degree.
For example, \citet{brown-2020-gpt3} demonstrate strong performance of LMs on multi-hop reasoning tasks such as DROP \cite{dua-2019-drop} which requires discrete reasoning and numeracy.
On the other hand, many argue that LM-based QA models are not actually performing any reasoning, and rather performing (sophisticated) pattern matching and data memorization \cite{marcus-2020-gpt}.
Simultaneously, in the context of extractive QA models that select answers from the provided context, several works have demonstrated that they can leverage superficial signals to return correct answers even when the context does not contain all the supporting facts \cite{chen-2019-multi,min-2019-multi}

In this paper, we perform a closer examination of the multi-hop reasoning capabilities of generative QA models.
To do so, we take multi-hop questions and their component single-hop questions to directly query generative QA models, studying their multi-hop reasoning ability.
Specifically, we use multi-hop questions from the ComplexWebQuestions \cite{talmor-2018-cwq} and HotpotQA \cite{yang-2018-hotpotqa,tang-2021-hotpotqa_dec} datasets as the testbed, and generate decomposed single-hop questions using heuristics (\autoref{sec:genqa_data}).
We examine two types of generative QA models, namely \emph{closed-book} \cite{roberts-closet5-2020,khashabi-2020-unifiedqa} and \emph{open-book} \cite{guu-2020-realm,lewis-2020-rag,izacard-2021-fid,xiong-2020-mdr} QA models that either do not or do refer to external knowledge when generating the answer respectively.
Specifically, we use UnifiedQA \cite{khashabi-2020-unifiedqa} as a representative closed-book model, and RAG \cite{lewis-2020-rag} as a representative open-book model (\autoref{sec:genqa_model}).
We first ask:
\begin{description}[itemsep=-1pt]
\item[RQ1] Is the correctness of decomposed single-hop questions a necessary and sufficient condition for correctness of multi-hop questions? (\autoref{sec:confusion}) Are answers to multi-hop and chains of decomposed questions consistent? (\autoref{sec:consist})
\item[RQ2] Do models trained on single-hop questions demonstrate zero-shot generalization to multi-hop questions? (\autoref{sec:zeroshot})
\end{description}
We find that generative QA models, even those close to the state-of-the-art, generally do not demonstrate robust multi-hop reasoning abilities, with success on multi-hop questions largely a result of taking shortcuts rather than true multi-hop reasoning.
Zero-shot multi-hop reasoning ability does not emerge naturally from training on single-hop questions, which motivates our final question:
\begin{description}[itemsep=-1pt]
\item[RQ3] Can we improve models' zero-shot multi-hop reasoning capacity by training on approximations of real multi-hop questions? (\autoref{sec:zeroshot})
\end{description}
Motivated by the fact that pre-training on massive text endows LMs with the ability to identify semantically similar expressions, our first method uses concatenated decomposed single-hop questions to approximate real multi-hop questions.
Our second method is inspired by recent work teaching LMs complex reasoning capabilities through neural execution of logical forms, e.g.~by training neural models to execute SQL queries \cite{liu-2021-tapex}.
We hypothesize that the ability to perform multi-hop reasoning can also be potentially learned from logical forms without reliance on NL questions.
To this end, we propose to use SPARQL, a standard query language over knowledge bases, as our logical forms to endow generative QA models with the ability to perform multi-hop reasoning, and examine whether learning to execute SPARQL transfers to the ability to answer NL multi-hop questions.
Both methods lead to significant improvement on zero-shot multi-hop reasoning performance, and further improvements are obtained when both are combined, opening possibilities for future work (\autoref{sec:future}).

\section{Generative Question Answering}
\label{sec:genqa}
In this section, we briefly introduce generative QA models and multi-hop QA datasets.
Then we elaborate on how we use multi-hop and decomposed questions to perform experiments.

\subsection{Generative QA Models}
\label{sec:genqa_model}
There are two main classes of generative QA models: closed-book and open-book.
Closed-book QA models usually consist of a sequence-to-sequence model that takes in a question $\bm{q}$ and calculates the probability of an answer $\bm{a}$ based on model parameters $\theta$ \cite{roberts-closet5-2020,khashabi-2020-unifiedqa}:
\begin{equation*}\compacteq
P(\bm{a}|\bm{q};\theta) = \prod_{i=1}^{|\bm{a}|}{P(a_i| \bm{q}, \bm{a}_{<i}};\theta),
\end{equation*}
Because these models can only refer to model parameters, any relevant information must be stored in the parameters \cite{roberts-closet5-2020}.
Open-book QA models first retrieve relevant context $\bm{c}$ from external resources, then generate answers using both questions and context \cite{guu-2020-realm,lewis-2020-rag,izacard-2021-fid}:
\begin{equation*}\compacteq
P(\bm{a}|\bm{c}, \bm{q};\theta) = \prod_{i=1}^{|\bm{a}|}{P(a_i|\bm{c}, \bm{q}, \bm{a}_{<i}};\theta),
\end{equation*}
We examine both types of models since we hypothesize that the difference in inputs might lead to different mechanisms of multi-hop reasoning.

\begin{table*}
\scriptsize
\centering
\vspace{-4mm}
\begin{tabular}{@{}l@{\tinycol}l@{}}
\toprule
\textbf{Type} & \textbf{Questions (hop1, hop2, and multi-hop)} \toright{\textit{\textbf{Answers}}} \\
\midrule
\multirow{3}{*}{Composition} & Return the country where Limonese Creole is spoken. \toright{\textit{Costa Rica}} \\
 & Which continent is \underline{Costa Rica} located? \toright{\textit{North America}} \\
 & On which continent is Limonese Creole spoken? \toright{\textit{North America}} \\
\midrule
\multirow{3}{*}{Conjunction} & What team is Reggie Bush on 2011? \toright{\textit{Miami Dolphins, New Orleans Saints}} \\
 & Which one of the following is the team won the super bowl XLIV championship: \underline{Miami Dolphins, New Orleans Saints}? \qquad\toright{\textit{New Orleans Saints}} \\
 & What team that won the super bowl XLIV championship was Reggie Bush in 2011? \toright{\textit{New Orleans Saints}} \\
\midrule
\multirow{3}{*}{Superlative} & What countries does the Niger River flow through? \toright{\textit{Benin, Guinea, Mali, Niger Nigeria}} \\
 & Which one of the following country calling code is smallest: \underline{Benin, Guinea, Mali, Niger, Nigeria}? \toright{\textit{Mali}} \\
 & What country with the smallest calling code does the Niger River flow through? \toright{\textit{Mali}} \\
\midrule
\multirow{3}{*}{Comparative} & What were Hitler's parents names? \toright{\textit{Alois Hitler, Klara Hitler}} \\
 & Which one of the following person's date of death is after 1903-01-03: \underline{Alois Hitler, Klara Hitler}? \toright{\textit{Klara Hitler}} \\
 & Which of Hitler's parents died after 3 January 1903? \toright{\textit{Klara Hitler}} \\
\bottomrule
\end{tabular}
\caption{Each multi-hop question $\bm{q}$ from ComplexWebQuestions is decomposed into two single-hop questions $\bm{q}_1$ and $\bm{q}_2$. \underline{Underlined entities} in the second single-hop questions are actually the answer to the first hop.}
\label{tab:cwq}
\end{table*}

Specifically, as our example of a closed-book model we use the UnifiedQA model of \citet{khashabi-2020-unifiedqa}.
The UnifiedQA model is based on the T5 model \cite{raffel-2020-t5}, which is an encoder-decoder model trained on the Colossal Clean Crawled Corpus (C4) by a denoising objective.
It further fine-tunes on a variety of QA datasets by converting different QA formats into a unified sequence-to-sequence format.

We use the RAG model of \citet{lewis-2020-rag} as our example of an open-book QA model, which consists of a retriever for searching relevant passages $\bm{p}$, and a generator which generates answers $\bm{a}$ given both $\bm{p}$ and $\bm{q}$.
The retriever is based on the dense passage retrieval model (DPR) \cite{karpukhin-2020-dpr}, and the generator is based on BART \cite{lewis-2020-bart}, which is also an encoder-decoder model that encodes both context and question, and generates answers autoregressively.

\subsection{Multi-hop Questions and Decompositions}
\label{sec:genqa_data}
To understand multi-hop reasoning in generative QA models, we propose to query models using both multi-hop questions and their decompositions into multiple single-hop questions, and perform analysis based on the predictions.

To this end, we choose the \textbf{ComplexWebQuestions} dataset \cite{talmor-2018-cwq} as our major testbed, as it contains multi-hop questions based on simple questions from the WebQuestionsSP dataset \cite{yin-2016-wq}, and we can leverage simple heuristics to obtain decomposed single-hop questions and corresponding answers.
Another advantage of ComplexWebQuestions is that it contains four types of questions: composition, conjunction, superlative, and comparative.
This allows us to perform fine-grained analysis over these categories.
Specifically, we follow heuristics in \citet{talmor-2018-cwq} to generate decompositions.
For the composition type, they use questions from WebQuestionsSP as the second hop, and replace an entity in it with a relational phrase to generate multi-hop questions.
We revert this process to get the first-hop question.
For the other three types, they use questions from WebQuestionsSP with multiple answers as the first hop, and add additional conditions to form the multi-hop questions.
We extract those conditions and use the following template to generate the second hop question: ``Which one of the following [condition]: [candidate answers]''.
\autoref{tab:cwq} includes examples of multi-hop questions and their decompositions of four types.

We also use another small dataset from \citet{tang-2021-hotpotqa_dec} to test the generality of models, where a subset of multi-hop questions from \textbf{HotpotQA} \cite{yang-2018-hotpotqa} are manually annotated with decompositions.
This dataset only contains a single type of question, which is composition.
ComplexWebQuestions has 27,639/3,519 questions in the training/development set, and HotpotQA has 1,000 questions in the development set.\footnote{Since the test sets of both datasets are hidden, we use development sets for evaluation purposes. 
Break \cite{wolfson-break-2020} is another testbed with multi-hop questions and manually decomposed questions.
However, the decomposed questions are not annotated with answers, making it less appropriate for our study.}

\subsection{Answer Generation and Evaluation}\label{sec:ans}
We use $\bm{q}_t, t \in \{1, ..., T\}$ to denote the $t$-th decomposed single-hop question for a multi-hop question $\bm{q}$ with $T$ hops.
Correspondingly, we use $\bm{a}_t$ to denote answers and $\bm{c}_t$ to denote retrieved context for the single-hop question $\bm{q}_t$.
Since the last single-hop question always has the same answer as the corresponding multi-hop question, $\bm{a}_T = \bm{a}$.
We use $\hat{\bm{a}}_t$/$\hat{\bm{a}}$ to denote the predictions from single-/multi-hop questions generated with greedy decoding:
\begin{equation*}
\tworow{\hat{\bm{a}}}{\hat{\bm{a}_t}} = \arg\max_{\bm{y}} P\Big(\bm{y}\Big\vert\tworow{[\bm{c},] \bm{q}}{[\bm{c}_t,] \bm{q}_t}; \theta\Big).
\end{equation*}
We query models using all decomposed questions $\bm{q}_t$ and multi-hop questions $\bm{q}$ which are concatenated with the corresponding context ($\bm{c}_t$ or $\bm{c}$) for open-book settings to get predicted answers.
All questions from ComplexWebQuestions and HotpotQA have two hops (i.e., $T=2$), thus in the following sections we always use $T=2$.

\paragraph{Pseudo-gold context for \openoracle models}
Previous work clearly demonstrates that a better retrieval component usually implies higher open-book QA performance, as it results in more retrieved contexts with answers \cite{chen-2017-drqa,lee-2019-orqa,karpukhin-2020-dpr}.
Therefore, we ablate out the influence of the retrieval component and focus on understanding the mechanism through which generative QA models parse multi-hop questions and generate answers.

We try to provide context that contains answers to the QA model so failure of answering the question can be mainly attributed to the generator instead of the retriever.
Since gold context is not annotated in the datasets, we follow \citet{karpukhin-2020-dpr} to obtain pseudo-gold context.
Specifically, we use the DPR model to retrieve the top-100 passages to each single-hop question $\bm{q}_t$, and find the first one containing the answer $\bm{a}_t$, which is denoted as the pseudo-gold passage $\bm{p}_t^{\cmark}$.
Only using pseudo-gold passages as the context might make the task too easy because no incorrect contexts are presented.
Therefore, we concatenate the pseudo-gold passage with a negative passage $\bm{p}_t^{\xmark}$ which is the first retrieved passage not containing the answers: $\bm{c}_t = [\bm{p}_t^{\cmark}, \bm{p}_t^{\xmark}]$.\footnote{The concatenation order is randomized to avoid leaking superficial signals to QA models.}
For multi-hop questions $\bm{q}$, we concatenate all context of the decomposed questions: $\bm{c} = [\bm{c}_1, ..., \bm{c}_T]$.
We fix the context for all of our experiments, and only use the generator of the RAG model.
For clarity, instead of open-book we use \emph{\openoracle} to refer to these QA models in the following sections.

\paragraph{Multi-answer generation}
Since some questions involve multiple answers, as shown in \autoref{tab:cwq}, we fine-tune generative QA models to generate multiple answers separated by a special symbol ``\#''.

\paragraph{Evaluation metrics}
We follow previous works \cite{roberts-closet5-2020,khashabi-2020-unifiedqa,lewis-2020-rag} to use exact match (EM) as our major evaluation metric, which measures the percentage of predictions that match the ground truth answers exactly \cite{rajpurkar-2016-squad,yang-2018-hotpotqa}.
Since we allow multi-answer generation, we split the prediction by the special symbol ``\#'' and match each entry against all the answers.
The prediction is judged as correct if all answers are included and no extra entry is predicted.

\section{Probing Multi-hop Questions and Decompositions}\label{sec:con}

To answer the first research question, we probe generative QA models on both multi-hop questions and their decompositions, examining the similarities and differences in models' behavior thereon.
We hypothesize that if models answer multi-hop questions in a robust way, they should be able to perform multi-hop reasoning by following the chain of decompositions internally, which makes being able to answer decomposed questions a necessary and/or sufficient condition of being able to answer multi-hop questions.
Motivated by this, we choose two probing angles to examine this question.
The first angle evaluates the prediction correctness on decomposed and multi-hop questions, and investigates whether there is a correlation between them.
The second angle generates predictions by answering multi-hop questions and the corresponding chain of decomposed single-hop questions in a sequence, and examining whether predictions are consistent.

\begin{table}[t]
\small
\centering
\begin{tabular}{ll|rrr}
\toprule
\textbf{Model} & \textbf{Type} & \textbf{Hop1} & \textbf{Hop2} & \textbf{Multi-hop} \\
\midrule
\multirow{5}{*}{\rotatebox[origin=c]{90}{UnifiedQA}} & overall & 32.91 & 49.13 & 33.25 \\
\cmidrule{2-5} 
& composition & 47.49 & 38.67 & 33.40 \\
& conjunction & 22.49 & 63.30 & 38.01 \\
& superlative & 16.23 & 48.69 & 21.99 \\
& comparative & 15.53 & 25.57 & 8.68 \\
\midrule
\multirow{5}{*}{\rotatebox[origin=c]{90}{RAG}} & overall & 58.72 & 65.11 & 60.32 \\
\cmidrule{2-5} 
& composition & 76.23 & 61.24 & 60.51 \\
& conjunction & 25.12 & 78.82 & 66.50 \\
& superlative & 13.33 & 76.67 & 53.33 \\
& comparative & 17.65 & 35.29 & 26.47 \\
\bottomrule
\end{tabular}
\caption{EM of two models on ComplexWebQuestions overall or each type separately.}
\label{tab:overall_cwq}
\end{table}

\begin{table}[t]
\small
\centering
\begin{tabular}{l@{\tinycol}l|rrr}
\toprule
\textbf{Model} & \textbf{Type} & \textbf{Hop1} & \textbf{Hop2} & \textbf{Multi-hop} \\
\midrule
UnifiedQA & composition & 1.70 & 1.30 & 1.20 \\
RAG & composition & 31.55 & 21.66 & 6.15 \\
\bottomrule
\end{tabular}
\caption{EM of two models on HotpotQA.}
\label{tab:overall_hp}
\end{table}

\begin{figure*}[ht]
\centering
\vspace{-4mm}
\includegraphics[width=1\textwidth, clip, keepaspectratio]{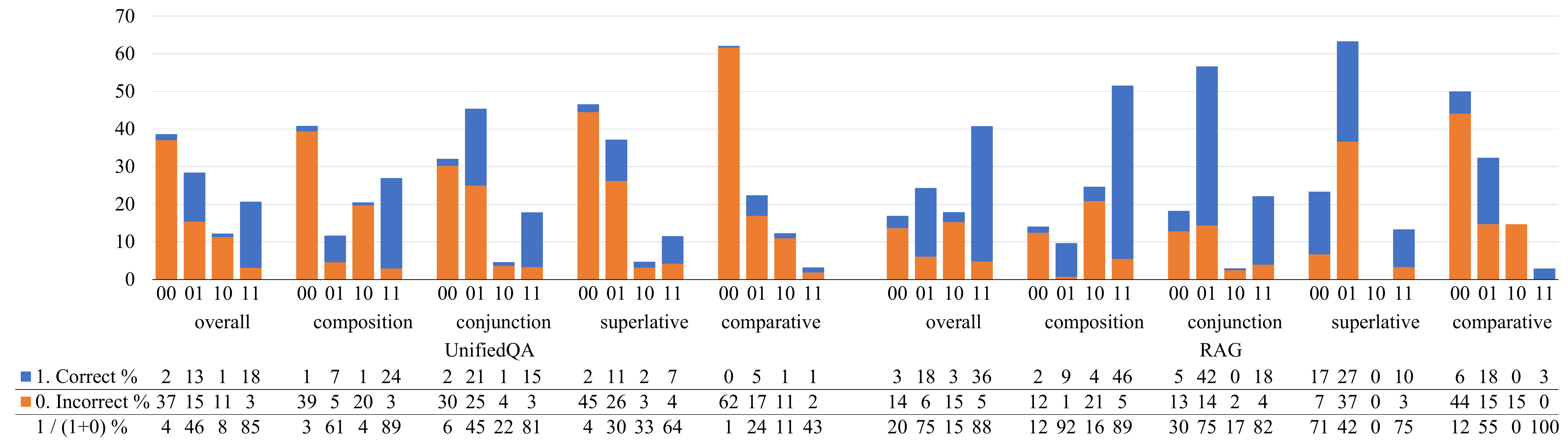}
\caption{Correctness confusion matrices of two models on ComplexWebQuestions. Two binary codes on the X-axis indicates the correctness of the first/second single-hop question $s_1s_2=\{00,01,10,11\}$. In the table, the first/second row indicates the percentage (\%) of examples of which the multi-hop question is correctly/incorrectly answered $P(s=\{1,0\}, s_1s_2)$; the last row indicates the conditional success rate $P(s=1|s_1s_2)$.}
\label{fig:confusion}
\end{figure*}

\subsection{Experimental Settings}
We fine-tune the UnifiedQA and RAG model using both single- and multi-hop QA pairs from the training set of the ComplexWebQuestions dataset.%
\footnote{We follow the default hyperparameters of UnifiedQA for 100K steps and a batch size of 16 on a single TPU, and the default hyperparameters of RAG for 10 epochs with a batch size of 4 on a single V100 GPU.}
Then we generate predictions for both single- and multi-hop questions from the test set of the ComplexWebQuestions/HotpotQA datasets, and show their overall results in \autoref{tab:overall_cwq} and \autoref{tab:overall_hp} respectively.
We measure the EM metric on first-hop $\bm{q}_1$ (\textbf{Hop1}), second-hop $\bm{q}_2$ (\textbf{Hop2}), and multi-hop questions $\bm{q}$ (\textbf{Multi-hop}) separately.
We also group examples by four types to investigate whether different types of reasoning exhibit different regularities.

To examine the correlation between success on decomposed and multi-hop questions, we bucket examples by their correctness.
We use $s_1$, $s_2$ and $s$ to denote correctness of predictions generated from the first/second single-hop and multi-hop questions, which is either 0 (incorrect) or 1 (correct).
There are $8=2^3$ \emph{configurations} of the correctness of a triple.
We present the results using the correctness confusion matrices in \autoref{fig:confusion}, where examples are bucketed into 4 bins by correctness on single-hop questions (i.e., $s_1 s_2=\{00, 01, 10, 11\}$) and the inner blue/orange bars indicate the percentage of the corresponding configurations (i.e., $P(s=\{1,0\}, s_1s_2=\{00,01,10,11\})$).
To better reveal the correlation between decomposed and multi-hop questions, we compute the \emph{conditional success rate} on multi-hop questions $P(s=1|s_1s_2)=\frac{P(s=1,s_1s_2)}{P(s=1,s_1s_2)+P(s=0,s_1s_2)}$ in the last row of the table, which indicates how likely multi-hop questions are correctly answered given the correctness on single-hop decompositions.%
\footnote{For robust models, $P(s=1|s_1s_2=11)$ should be close to 1, $P(s=1|s_1s_2=\{00,01,10\})$ should be close to 0.}

To examine the prediction consistency between multi-hop questions and chains of decompositions, we replace entities in the second single-hop questions $\bm{q}_2$ which correspond to answers to the first hop with a special placeholder ``\#1'', and denoted it as $\bm{q}_2^*$.
When answering a chain of decomposed questions, predictions from the first hop $\hat{\bm{a}}_1$ are used to replace the placeholder in the second hop: $\bm{q}_2^*(\hat{\bm{a}}_1)$, from which we generate the final answer denoted as $\hat{\bm{a}}_2^*$.
Models fine-tuned in the normal setting only generate final answers from multi-hop questions, but not intermediate answers (i.e., answers to the hop1 question).
To examine whether models can predict intermediate answers from the multi-hop question, and measure consistency on both, we append two prompts to multi-hop questions to instruct models to generate two predictions:

\vspace{-2mm}
{\small
\begin{align*}\compacteq
\tworow{\hat{\bm{a}}_1^*}{\hat{\bm{a}}} &= \arg\max_{\bm{y}} P\Big(\bm{y}\Big\vert[\bm{c},] \bm{q}, \tworow{\text{``Intermediate answer:''}}{\text{``Final answer:''}}\Big) \\
\tworow{\hat{\bm{a}}_1}{\hat{\bm{a}}_2^*} &= \arg\max_{\bm{y}} P\Big(\bm{y}\Big\vert\tworow{[\bm{c}_1,] \bm{q}_1}{[\bm{c}_2,] \bm{q}_2^*(\hat{\bm{a}}_1)}\Big),
\end{align*}
}%
where $\hat{\bm{a}}_1^*$ denotes intermediate predictions.
Predictions from multi-hop questions ($\hat{\bm{a}}_1^*$/$\hat{\bm{a}}$) are compared with predictions from decomposed questions in sequence ($\hat{\bm{a}}_1$/$\hat{\bm{a}}_2^*$) respectively to measure their consistency.

\subsection{Correlation of Correctness}
\label{sec:confusion}

\paragraph{Multi-hop performance is unexpectedly high}
Given the hypothesis that being able to answer decomposed questions is a prerequisite of being able to answer multi-hop questions, we expect \textit{a priori} that the performance on multi-hop questions will be much lower than the performance on all single-hop questions due to error propagation.
However, what we observe on ComplexWebQuestions is the opposite: overall, the multi-hop performance is slightly higher than the hop1 performance, and the gap between hop2 and multi-hop performance is much smaller than may be expected, especially for the \openoracle RAG model.
This indicates that generative QA models somehow manage to take shortcuts when answering multi-hop questions, i.e., being able to answer the multi-hop question without correctly answering its component questions.

\begin{table*}[ht]
\scriptsize
\vspace{-4mm}
\centering
\begin{tabular}{@{}c@{\thincol}p{0.6\textwidth}@{\tinycol}l@{}}
\toprule
\textbf{Type} & \textbf{Questions (hop1, hop2, multi-hop)} & \textbf{Answers} \toright{\textbf{Predictions}} \\
\midrule
\multirow{3}{*}{\rotatebox[origin=c]{90}{Compo.}} & Return the country where Cerveceria Modelo Corona light beer is made. & Mexico \toright{\correct{Mexico}} \\
 & Who is Mexico's president right now 2011? & Felipe Calderón \toright{\correct{Felipe Calderón}} \\
 & Who was the president in 2011 in the country where Cerveceria Modelo Corona light beer is made? & Felipe Calderón \toright{\wrong{Juan Manuel Santos}} \\
\midrule
\multirow{3}{*}{\rotatebox[origin=c]{90}{Conj.}} & What year did Detroit Pistons win the championship? & 2004, 1990, 1989 NBA Finals \toright{\correct{2004, 1990, 1989 NBA Finals}} \\
 & Which one of the following sports championship results were 4-1: 2004, 1990, 1989 NBA Finals? & 2004, 1990 NBA Finals \toright{\correct{2004, 1990 NBA Finals}} \\
 & In what year did the Detroit Pistons win the sports championship where the results were 4-1? & 2004, 1990 NBA Finals \toright{\wrong{2002 NBA Finals}} \\
\midrule
\multirow{3}{*}{\rotatebox[origin=c]{90}{Compo.}} & Return the team won the 2006 NFC championship & Seattle Seahawks \toright{\wrong{Indianapolis Colts}} \\
 & Where do the Seattle Seahawks play? & CenturyLink Field \toright{\correct{CenturyLink Field}} \\
 & Which Stadium does the team that claimed the 2006 NFC championship play in? & CenturyLink Field \toright{\correct{CenturyLink Field}} \\
\midrule
\multirow{3}{*}{\rotatebox[origin=c]{90}{Compa.}} & Who is the leader of France 2012? & Nicolas Sarkozy, François Hollande \toright{\wrong{Nicolas sarkozy}} \\
 & Which one of the following started tenure after 1979: Nicolas Sarkozy, François Hollande? & Nicolas Sarkozy \toright{\correct{Nicolas Sarkozy}} \\
 & Who was the leader of France from 1979 until 2012? & Nicolas Sarkozy \toright{\correct{Nicolas Sarkozy}} \\
\bottomrule
\end{tabular}
\caption{Cases of predictions generated from single/multi-hop questions of different types. Correct/Incorrect predictions are indicated in \correct{blue}/\wrong{orange}.}
\label{tab:case}
\end{table*}

\paragraph{Success on decompositions does not always imply success on multi-hop questions}
Looking at the overall percentage, we can see that indeed the success rate on multi-hop questions is highest if both decomposed questions are correctly answered, i.e., $P(s=1|s_1s_2=11)=85\%/88\%$ for the UnifiedQA/RAG model respectively, indicating that generative QA models are more likely to answer multi-hop questions if they can answer all decomposed single-hop questions.
However, there are still 15\%/12\% examples where correctness on both decomposed questions does not imply correctness on multi-hop questions, as shown by the first two examples in \autoref{tab:case}.
The predictions generated from the multi-hop questions are usually of the correct type, but they diverge from the predictions generated from decomposed questions, indicating that models do not necessarily follow the decomposed components when answering multi-hop questions.

\paragraph{Multi-hop success is most correlated with success on the last hop}
Even when models fail on decomposed questions, they can still answer some percentage of multi-hop questions correctly (4-46\%/15-75\%) depending on which of the decomposed hops fails.
Success on hop2 questions is more correlated with success on multi-hop questions than hop1 questions (i.e., $P(s=1|s_1s_2=01) > P(s=1|s_1s_2=10)$), especially for the \openoracle RAG model.
When the model is only able to answer the second single-hop questions, there is still $46\%/75\%$ chance that the model can answer the multi-hop questions in closed/\openoracle settings respectively, indicating that generative QA models manage to take shortcuts instead of performing real reasoning.
The shortcuts could be some superficial signals in the context or parameters that generative QA models can take advantage of to bypass the requirement of the first hop, as shown by the third example in \autoref{tab:case}.
Or for multi-hop questions with multiple intermediate answers, generative QA models might not need to know all of them in order to answer the multi-hop questions, as shown by the fourth example in \autoref{tab:case}.

\paragraph{Other observations}
Overall, the \openoracle RAG model performs significantly better than the closed-book UnifiedQA model on both datasets, indicating that knowledge stored in parametric generative QA models is still limited and it is beneficial to provide external evidence.
Hop2 performance is significantly higher than the hop1 performance on conjunction, superlative, and comparative questions, which is because hop1 questions usually have more answers than hop2 questions as shown in \autoref{tab:cwq}, thus being harder.%
\footnote{Note that the difference in the difficulty of hop1 and hop2 questions does \emph{not} invalidate our previous conclusion about correctness correlation since we use conditional success rate.}
Both models generalize poorly to the unseen HotpotQA dataset (\autoref{tab:overall_hp}), indicating that the learned multi-hop reasoning capability cannot generalize across datasets.

\subsection{Prediction Consistency}
\label{sec:consist}

\begin{table}
\small
\vspace{-2mm}
\centering
\begin{tabular}{@{}l@{\tinycol}lr@{\tinycol}rr@{\tinycol}r|r@{\tinycol}r}
\toprule
 & \textbf{Type} & \multicolumn{4}{c|}{\textbf{EM}} & \multicolumn{2}{c}{\textbf{Consistency}} \\
 &  & \multicolumn{2}{c}{\textbf{Decompose}} & \multicolumn{2}{c|}{\textbf{Multi-hop}} &  \\
 &  & \textbf{Hop1} & \textbf{Hop2} & \textbf{Hop1} & \textbf{Hop2} & \textbf{Hop1} & \textbf{Hop2} \\
\midrule
\multirow{5}{*}{\rotatebox[origin=c]{90}{UnifiedQA}} & overall & 32.48 & 32.23 & 30.78 & 31.40 & 50.81 & 36.12 \\
\cmidrule{2-8} 
 & compo. & 51.87 & 33.97 & 48.51 & 32.13 & 58.92 & 43.24 \\
 & conj. & 17.73 & 34.68 & 17.54 & 34.68 & 44.46 & 33.51 \\
 & super. & 13.09 & 24.08 & 11.52 & 23.04 & 38.74 & 26.70 \\
 & compa. & 13.24 & 9.59 & 12.79 & 10.50 & 47.49 & 11.42 \\
\midrule
\multirow{5}{*}{\rotatebox[origin=c]{90}{RAG}} & overall & 56.51 & 62.65 & 61.92 & 58.11 & 79.61 & 65.48 \\
\cmidrule{2-8}
 & compo. & 73.86 & 60.88 & 76.78 & 54.48 & 86.47 & 67.46 \\
 & conj. & 23.65 & 74.38 & 30.05 & 72.41 & 67.98 & 68.47 \\
 & super. & 13.33 & 60.00 & 33.33 & 56.67 & 60.00 & 43.33 \\
 & compa. & 11.76 & 23.53 & 38.24 & 32.35 & 55.88 & 35.29 \\
\bottomrule
\end{tabular}
\caption{EM of predictions from answering chains of decomposed questions and multi-hop questions on ComplexWebQuestions and their consistency (\%).}
\label{tab:consist}
\vspace{-3mm}
\end{table}

\paragraph{Predictions are not consistent between multi-hop questions and chains of decompositions}
As shown in \autoref{tab:consist}, consistency is relatively low for both models and on both hop1 and hop2, indicating that generative QA models answer multi-hop questions not necessarily in the same way as they answer decomposed questions in sequence.
The consistency of the UnifiedQA model is lower than the consistency of the RAG model, which is because knowledge is limited in closed-book QA models, and navigating in parameters implicitly is probably harder than searching chains of evidence in context explicitly.
Consistency on the first hop is usually higher than consistency on the second hop, which is because inconsistent intermediate predictions ($\hat{\bm{a}}_1$) will propagate to the second hop, leading to accumulated inconsistency.

\section{Improving Zero-shot Multi-hop Reasoning Capability}
\label{sec:zeroshot}

In this section, we first examine LMs' zero-shot capacity for multi-hop reasoning when they are \emph{not} trained on multi-hop NL questions.
Compositional generalization ability \cite{lake-2018-comp} is required in this case to generalize from single-hop to multi-hop questions.
Unsurprisingly, generative QA models perform poorly in this setting, with almost half performance degradation.
Since multi-hop NL questions are expensive to obtain, one natural question is ``is it possible to improve the multi-hop reasoning ability using only single-hop NL questions, or even without any NL questions?''

\begin{figure}[tb]
\includegraphics[width=1.0\columnwidth, clip, keepaspectratio]{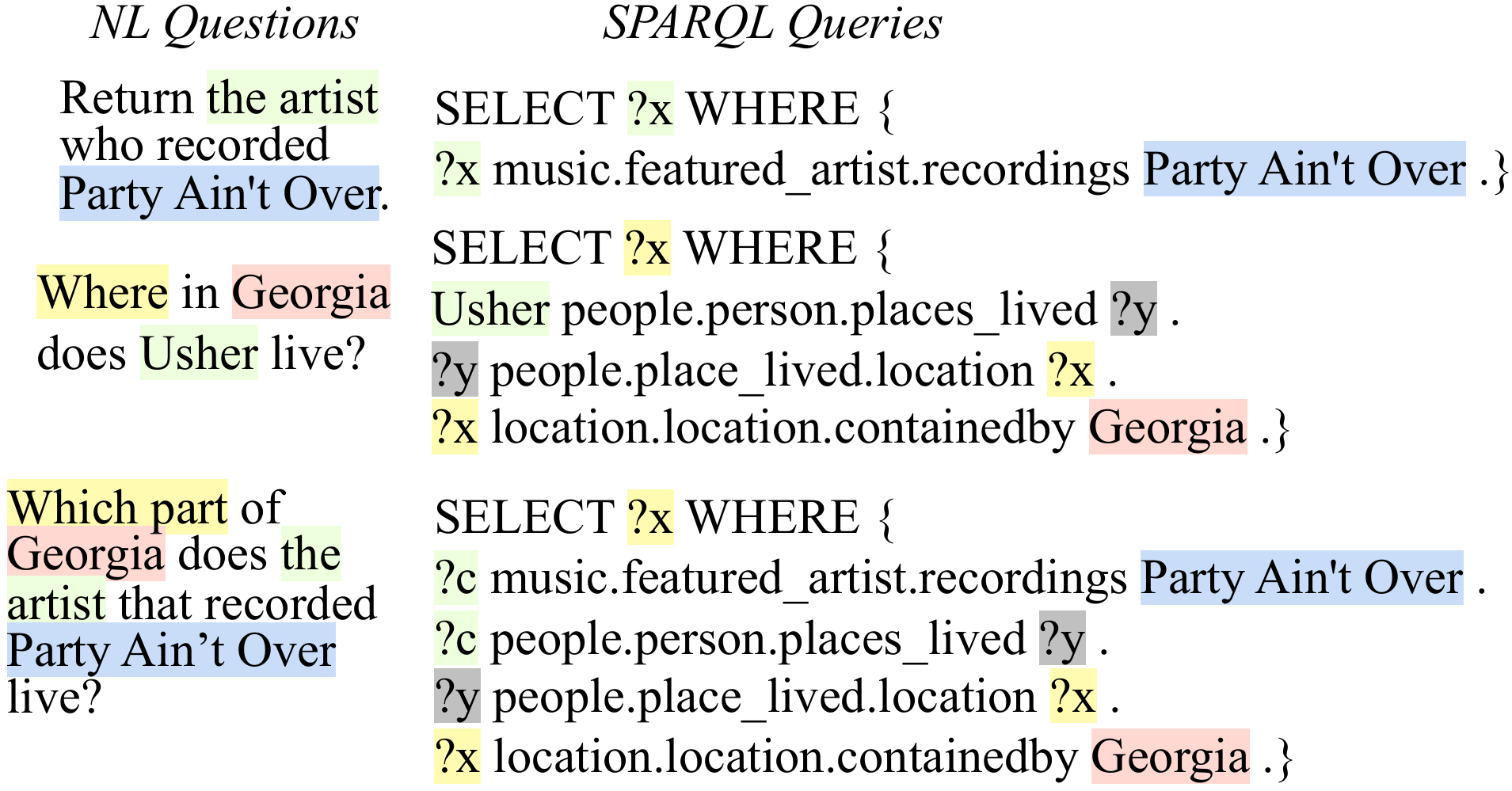}
\centering
\caption{NL questions and corresponding SPARQL queries. Mentions of the same entity are in the same color.}
\label{fig:sparql}
\end{figure}

We design two methods to achieve this goal.
Motivated by the fact that UnifiedQA and RAG models are initialized with language models pre-trained on massive text corpora, which endows them with the ability to identify semantically similar expressions, our first method uses concatenated decomposed NL questions (i.e., $[\bm{q}_1, \bm{q}_2^*]$) to approximate the real NL multi-hop question $\bm{q}$, and fine-tunes models on them.

The second is inspired by recent progress in teaching LMs complex reasoning capabilities by executing logical forms neurally.
For example, \citet{liu-2021-tapex} formulate the execution of SQL over tables as a seq2seq task where the input is a logical form string associated with a table and the output is answers \cite{liu-2021-tapex}.
We hypothesize that in our multi-hop QA setting, the ability to perform multi-hop reasoning can also be potentially learned from logical forms without reliance on any NL question.
To this end, we propose to use SPARQL, which is a standard query language over knowledge bases, as our logical forms. We then examine whether the ability to answer questions expressed in these SPARQL queries is transferable to NL multi-hop questions.
The advantage of using SPARQL for training is that SPARQL queries can be easier and cheaper to obtain or generate than NL.
For example we can use existing query logs,\footnote{\url{https://bit.ly/3wRRIPZ}} or use manual SPARQL queries as templates and replace entities/relation to generate more queries.\footnote{\url{https://bit.ly/3ciOFqy}}
Our observation sheds light on potential improvement on multi-hop reasoning using many SPARQL query-answer pairs.

\subsection{Experimental Settings}
Each NL multi-hop question in the ComplexWebQuestions dataset is associated with a SPARQL query based on the Freebase schema.
We follow similar heuristics described in \autoref{sec:genqa_data} to generate SPARQL queries for the first- and second-hop NL questions.
Each single- and multi-hop SPARQL query is used as a pseudo input question after replacing entity identifiers with their names, as shown in \autoref{fig:sparql}.
In order to answer the above research questions, we design the following settings:
\begin{itemize}
\item No fine-tuning (\textbf{Default}): This setting uses the original model without fine-tuning.
\item \textbf{S}ingle- and \textbf{M}ulti-hop \textbf{NL} (\textbf{SM-NL}): The normal setting discussed in \autoref{sec:confusion} where we train the model using both single- and multi-hop NL questions.
This serves as the upper bound of the zero-shot performance.
\item \textbf{S}ingle-hop \textbf{NL} (\textbf{S-NL}): Only use decomposed single-hop NL questions for training.
\item \textbf{S}ingle- and \textbf{M}ulti-hop \textbf{SPARQL} (\textbf{SM-SPARQL}): Only use SPARQL queries.
\item \textbf{S-NL} with \textbf{Concat}enation (\textbf{S-NL+Concat}): Use concatenated decomposed NL questions in addition to \textbf{S-NL}.
\item \textbf{SM-NL+Concat+SM-SPARQL (Combo.)}: Combine the previous two settings to leverage both NL and SPARQL for training.
\end{itemize}

\newcolumntype{R}{@{}>{\columncolor{white}[0pt][0pt]}r@{}}
\begin{table}
\small
\centering
\begin{tabular}{l@{\smallcol}l@{\smallcol}c@{\smallcol}c@{\smallcol}r@{\smallcol}r@{\smallcol}r}
\toprule
 & \multirow{2}{*}{\textbf{Setting}} & \multicolumn{2}{c}{\textbf{Supervision}} & \multirow{2}{*}{\textbf{Hop1}} & \multirow{2}{*}{\textbf{Hop2}} & \textbf{Multi-} \\
 & & \textbf{Single} & \textbf{Multi} & & & \textbf{hop} \\
\midrule
\multirow{6}{*}{\rotatebox[origin=c]{90}{UnifiedQA}} & Default & & & 0.71 & 15.37 & 6.56 \\
& S-NL & \nl & & 33.28 & 49.33 & 17.02 \\
& \quad+Concat & \nl & \concat & 31.91 & 48.25 & 25.69 \\
& SM-SPARQL & \sparql & \sparql & 19.04 & 34.67 & 24.84 \\
& Combo. & \nl\sparql & \concat\sparql & 32.76 & 48.51 & \textbf{27.14} \\
& \hlcell SM-NL & \hlcell\nl & \hlcell\nl & \hlcell 32.91 & \hlcell 49.13 & \hlcell 33.25 \\
\midrule
\multirow{6}{*}{\rotatebox[origin=c]{90}{RAG}} & Default & & & 7.99 & 12.65 & 7.62 \\
& S-NL & \nl & & 59.83 & 68.55 & 34.03 \\
& \quad+Concat & \nl & \concat & 61.06 & 64.13 & \textbf{53.93} \\
& SM-SPARQL & \sparql & \sparql & 49.51 & 58.48 & 51.60 \\
& Combo. & \nl\sparql & \concat\sparql & 57.37 & 62.53 & 53.07 \\
& \hlcell SM-NL & \hlcell\nl & \hlcell\nl & \hlcell 58.72 & \hlcell 65.11 & \hlcell 60.32 \\
\bottomrule
\end{tabular}
\caption{EM on NL questions in zero-shot multi-hop evaluation, where \nl, \concat, \sparql denotes NL, concatenation, and SPARQL respectively. Oracle performance using multi-hop NL questions has a gray background. Best zero-shot multi-hop performance is in bold.}
\label{tab:zero}
\end{table}

\subsection{Experimental Results}
\autoref{tab:zero} includes results for all the above experimental settings.
Compared to the oracle multi-hop performance, performance of only using single-hop NL questions (\textbf{S-NL}) drops by almost half on both UnifiedQA ($33.25\veryshortarrow17.02$) and RAG models ($60.32\veryshortarrow34.03$), indicating that without learning on multi-hop questions, compositional generalization does not naturally emerge in generative QA.

\paragraph{Single-hop concatenation is a good approximation of multi-hop questions}
Surprisingly, by simply concatenating single-hop NL questions and fine-tuning on them, multi-hop performance increases by a large margin ($17.02\veryshortarrow25.69$/$34.03\veryshortarrow53.93$), indicating that simple concatenation is an effective approximation for multi-hop questions.
We hypothesize that LMs pre-trained on noisy text have the paraphrasing ability to generalize from concatenated simple sentences to complex sentences at least to some degree.

\paragraph{Models generalize from SPARQL to NL questions}
SPARQL queries explicitly specify compositional structure using pre-defined grammar and canonicalized entities/relations, while NL questions express this process in a more flexible way.
Despite this gap, models trained solely on SPARQL queries are able to generalize to NL questions at test time on both single- and multi-hop questions, with a performance drop of 7-15 on both single- and multi-hop questions compared to (\textbf{SM-SPARQL} vs. \textbf{SM-NL}), which is far better than no fine-tuning (\textbf{Default}).
This indicates that when answering NL questions, the ability learned from mapping SPARQL queries to answers can be reused by the model, similar to the observation on table-based QA \cite{liu-2021-tapex}.
As demonstrated in other tasks such as table-based QA \cite{jiang-2022-omnitab} and text-to-SQL \cite{wu-2021-data}, converting the SPARQL queries into NL questions and training models on them can potentially mitigate the gap and further improve the performance, which we plan to explore in future works.

\paragraph{Combining concatenation and SPARQL improves further}
In this setting, we attempt to combine the merits of using concatenated single-hop NL questions, which are more natural, and SPARQL queries, which are more explicit with respect to the reasoning process.
Compared to training on two types of supervision separately (\textbf{S-NL+Concat} and \textbf{SM-SPARQL}), training on both jointly (\textbf{Combo.}) improves the multi-hop performance of UnifiedQA (25.69$\veryshortarrow$27.14) while slightly hurting the performance of RAG (53.93$\veryshortarrow$53.07).
We hypothesize that closed-book models are less constrained compared to \openoracle models due to the existence of the additional context, therefore closed-book models can benefit from the stronger supervision from a combination of two methods.
Note that there is still a large gap between fine-tuning on multi-hop NL questions (\textbf{SM-NL}) and zero-shot settings, which indicates the potential for better approximations or modeling techniques.

\section{Related Work}
\paragraph{Multi-hop QA models}
Most multi-hop QA models proposed so far are pipeline methods that generate sub-questions to retrieve evidence iteratively \cite{qi-2019-golden,ding-2019-cogaq,qiu-2019-dfgn,das-2019-iterqa,asai-2020-pathretriever,min-2019-decomprc,perez-2020-unsupqa,xiong-2020-mdr}.
The final answers are generated either by reading each retrieved evidence independently and recomposing the generated intermediate answers \cite{min-2019-decomprc,perez-2020-unsupqa}, or by taking all evidence as input at once \cite{qi-2019-golden,das-2019-iterqa,asai-2020-pathretriever}.
Instead, we focus on understanding the multi-hop reasoning capabilities of end-to-end generative QA models in this paper.

\paragraph{Analysis of multi-hop reasoning}
Several works studying multi-hop reasoning in extractive QA models found that they can leverage superficial signals to extract answers even when the context does not contain all supporting facts \cite{chen-2019-multi,min-2019-multi,trivedi-2020-dire,jiang-bansal-2019-avoiding,niu-etal-2020-self,lee-etal-2021-robustifying}.
While they examine from the perspective of dataset bias, we directly query models with both multi-hop and component single-hop questions, using both closed- and open-book generative QA models.
Another work that studied multi-hop QA models using both multi-hop and single-hop questions is \citet{tang-2021-hotpotqa_dec}.
While they use pipeline extractive QA models, we focus on end-to-end generative QA models and investigate correctness, consistency, and compositional generalization ability.

\section{Conclusion}
\label{sec:future}
In this paper, we examined the multi-hop reasoning capabilities of generative QA models, finding that overall models take shortcuts when answering multi-hop questions, not demonstrating convincing multi-hop reasoning capability.
When trained only on single-hop questions, models generalize poorly to multi-hop questions, while approximation using the concatenation of single-hop questions and SPARQL queries improves the multi-hop performance significantly.
Further directions include better approximations of multi-hop questions and advanced modeling techniques that encourage compositional ability.

\bibliography{anthology,custom,acl_anthology_latest}
\bibliographystyle{acl_natbib}

\end{document}